\pdfoutput=1

\documentclass[11pt]{article}
\usepackage{custom}

\title{Challenges to Evaluating the Generalization of Coreference Resolution Models: A Measurement Modeling Perspective}

\author[1]{Ian Porada}
\author[2]{Alexandra Olteanu}
\author[*]{Kaheer Suleman}
\author[*]{Adam Trischler}
\author[1,3]{Jackie Chi Kit Cheung}
\affil[1]{Mila, McGill University}
\affil[ ]{\tt ian.porada@mail.mcgill.ca, jackie.cheung@mcgill.ca}
\affil[2]{Microsoft Research Montr\'eal}
\affil[ ]{\tt alexandra.olteanu@microsoft.com}
\affil[3]{Canada CIFAR AI Chair}

\begin{document}
\maketitle
\begin{abstract}
It is increasingly common to evaluate the same coreference resolution (CR) model on multiple datasets.
Do these multi-dataset evaluations allow us to draw meaningful conclusions about model generalization?
Or, do they rather reflect the idiosyncrasies of a particular experimental setup (e.g., the specific datasets used)?
To study this, we view evaluation through the lens of measurement modeling, a framework commonly used in the social sciences for analyzing the validity of measurements.
By taking this perspective, we show how multi-dataset evaluations risk conflating different factors concerning what, precisely, is being measured.
This in turn makes it difficult to draw more generalizable conclusions from these evaluations.
For instance, we show that across seven datasets, measurements intended to reflect CR model generalization are often correlated with differences in both how coreference is defined and how it is operationalized; this limits our ability to draw conclusions regarding the ability of CR models to generalize across any singular dimension.
We believe the measurement modeling framework provides the needed vocabulary for discussing challenges surrounding what is actually being measured by CR evaluations.
\end{abstract}

\def\thefootnote{*}\footnotetext[3]{Previously affiliated with Microsoft Research Montr\'eal.}
\renewcommand{\thefootnote}{\arabic{footnote}}

\section{Introduction}
\label{sec:introduction}

Coreference is the broad phenomenon of multiple linguistic expressions referring to the same discourse entity (see examples in Table~\ref{tab:coref-examples}).
Coreference resolution (CR) is then the task of identifying those expressions that are coreferring~\citep[][\textit{i.a.}]{winograd1972understanding,kantor1977management,hirst1981anaphora}.
This task has been studied extensively~\citep[\textit{e.g.,}][]{SUKTHANKER2020139,poesio2023computational} and is considered a core part of representing the semantics of natural language~\citep{hobbs1978resolving}.

\begin{table}[t]
    \centering
    \def\arraystretch{0.7}
    \scriptsize
    \begin{tabular}{@{}l  p{0.76\linewidth}@{}}
        Dataset & Example \\
        \midrule
        OntoNotes & Highway officials insist the ornamental railings on \hly{older bridges} aren't strong enough to prevent vehicles from crashing through. But other people don't want to lose \hly{the bridges'} beautiful, sometimes historic, features. \\[.75em]
        
        PreCo & ``Melting permafrost can also destroy \hlb{trees} and forests,'' Weller said. ``When holes in the ground form, \hlb{trees} fall into them and die.'' \\[-.25em]
        
        \makecell[l]{\\ Phrase \\ Detectives} & Within 1942-1944 \hlg{bicycles} were also added to regimental equipment pools. \ldots \ Although seeing heavy use in World War I, \hlg{bicycles} were largely superseded by motorized transport in more modern armies. \\
        \bottomrule
    \end{tabular}
    \vspace{-4pt}
    \caption{Examples of annotated coreference (highlighted) where at least one expression could be interpreted as generic. Generic expressions tend to be annotated differently across datasets in part due to varying definitions of coreference.}
    \label{tab:coref-examples}
    \vspace{-8pt}
\end{table}

As a result, a common goal has been to build CR models that are able to generalize across settings~\cite{urbizu-etal-2019-deep,xia-van-durme-2021-moving,zabokrtsky-etal-2022-findings}.
This goal has motivated the design of benchmarks ``to measure progress towards truly general-purpose coreference resolution''~\citep{toshniwal-etal-2021-generalization},
and has led to CR models being increasingly evaluated across multiple datasets~\citep[\textit{e.g.,}][]{yang-etal-2012-domain,poot-van-cranenburgh-2020-benchmark,straka-2023-ufal}. 

However, {\em what we can learn from multi-dataset evaluations about a CR model's ability to generalize will necessarily be limited by differences in how the datasets were constructed}. %

Consider the following example: researcher \textbf{A} defines coreference to be a relationship exclusively between noun phrases and uses crowdworkers to annotate a dataset based on this definition.
Independently, researcher \textbf{B} defines coreference to be a relationship between any noun phrase or verb phrase and annotates a different dataset based on this looser definition.
A CR model is then trained using \textbf{A}'s dataset and evaluated on \textbf{B}'s.
Without even seeing the evaluation results, one might already suspect that what we can learn from those results about how well a model trained on one dataset generalizes to the other will depend on the differences in how coreference was defined and annotated across the two datasets.

We aim to unpack how differences in the definition and annotation of coreference across datasets might limit the conclusions that can be drawn about the generalization ability of CR models.
To do so, we examine common evaluation practices using the {\em measurement modeling} framework~\citep[][\textit{i.a.}]{smelser_indicator_2001,jacobs2021measurement,blodgett-etal-2021-stereotyping}. 
This framework helps us distinguish between what is being measured, the \textit{construct}, from how it is being measured, the \textit{operationalization} of the construct via a \textit{measurement model}.
This distinction is particularly important when working with contested constructs---i.e., constructs with disputed or many competing definitions---which latent constructs often are, as it helps differentiate between differences in the definitions of a construct versus differences in operationalizations of the same definition.\looseness=-1

In fact, coreference is itself a contested construct~\citep{grishman-sundheim-1996-design,doddington-etal-2004-automatic,can_we_fix}; ``[a] very basic problem arising in the case of coreference is deciding what type of information is being annotated, since the term `coreference' is used to indicate different things''~\cite{poesio-etal-1999-mate}.
The measurement modeling framework thus helps us illustrate how common multi-dataset evaluations risk conflating different factors concerning what, precisely, is being measured; and, as a result, what we might be learning from these evaluations about models' ability to generalize across datasets.

In this work, we conduct a multi-dataset evaluation across seven English-language datasets of coreference annotations and analyze the results through the lens of measurement modeling.
Specifically, we consider how coreference is defined and annotated within each dataset~(\S\ref{sec:datasets}), and how this affects measurements of model generalization~(\S\ref{sec:experiments} \& \S\ref{sec:results}).
For this, we identify particular types of coreference that vary between definitions---e.g., the coreference of generic and predicative expressions---and show, using disaggregated results, how models systematically fail in cases where the definition of coreference used to build the test set differs from that used for the training set.

For instance, consider the coreference annotated in PreCo and highlighted in Table~\ref{tab:coref-examples}. A state-of-the-art CR model trained on OntoNotes fails to predict the two instances of ``trees'' as coreferring. This may be attributable to the fact that most often two generic nouns are not considered coreferring under the definition of coreference used in OntoNotes.

Our results suggest that to understand if and how CR models generalize some concept of coreference requires us to {\em first} resolve inconsistencies in how coreference is defined.
More broadly, through disaggregated evaluations and by applying the measurement modeling framework, our work provides a blueprint for evaluating out-of-domain performance in a way that accounts for known inconsistencies in how coreference is defined and operationalized across datasets.

\section{Background and Related Work}
\label{sec:related-work}

\paragraph{CR Model Error Analyses. }
Our work relates to efforts to identify and classify model errors.
Error analyses of CR models, however, have focused on performance measurements based on a single dataset~\citep{uryupina-2008-error,Versley2008,durrett-klein-2013-easy}.
Analyses that consider collections of datasets find that many types of errors are consistent across datasets, but often do not examine cases where models are tested on datasets they were not trained on~\citep{lu-ng-2020-conundrums,chai-strube-2023-investigating}.
Furthermore, these prior studies evaluating CR models' performance across datasets---as well as studies that analyze the models' downstream performance~\citep{dasigi-etal-2019-quoref,chai-etal-2022-evaluating}---have not considered how aspects related the way in which coreference is defined might correlate with the observed errors.

\paragraph{CR Model Generalization.} To improve generalization, several approaches aim to train a CR model using multiple training sets \citep[\textit{e.g.,}][]{zhao-ng-2014-domain,https://doi.org/10.48550/arxiv.2210.07602,yuan-etal-2022-adapting}. Such work has focused mostly on \textit{active learning}. It has been observed that models trained on multiple datasets do not improve in accuracy on a test set for which large amounts of in-domain training data is available~\citep{toshniwal-etal-2021-generalization}. Related to our work, \citet{moosavi-etal-2019-using} studied how coreference annotations differ in terms of the span boundaries annotated and proposed an algorithm for heuristically determining the minimal span boundary of a mention.

Models trained on OntoNotes are believed to be less accurate at CR for textual genres and for proper nouns not in the training set~\citep{moosavi-strube-2017-lexical,subramanian-roth-2019-improving,zhu-etal-2021-ontogum}. Prior work has improved on
existing generalization measurements 
in these cases by incorporating explicit linguistic features or syntactic rules~\citep{zeldes-zhang-2016-annotation,moosavi-strube-2018-using}.
Prior work has also examined the generalization of CR models to supposedly difficult cases of pronominal coreference~\citep{rahman-ng-2012-resolving,peng-etal-2015-solving,toshniwal-etal-2021-generalization}. For instance, models trained on OntoNotes are believed to generalize poorly to examples of pronominal coreference scraped from the web~\citep{webster-etal-2018-mind,emami-etal-2019-knowref}.

\paragraph{Measurement Modeling.}
\label{sec:measurement_modeling}
How to effectively measure theoretical concepts has been studied extensively in the quantitative social sciences~\citep[\textit{e.g.,}][]{black_doing_1999,adcock_collier_2001,bhattacherjee2012social}.
Because theoretical concepts are often unobservable and can therefore not be measured directly, they must be inferred based on observable variables or the inferred measurements of other theoretical concepts.
A model that describes the relationship between a theoretical concept and the variables from which that concept is inferred is called a \textit{measurement model}~\citep{smelser_indicator_2001,jacobs2021measurement}. 
Typically, the term \textit{construct} is used to refer to the theoretical concept being measured and the term \textit{indicator} is used to refer to any of the observable variables from which the measurement of the construct is inferred~\citep{kline_principles_2011}.\looseness=-1

We draw on these lines of work to examine how decisions made when constructing CR evaluation datasets might limit the type of conclusions we can draw about models' ability to generalize from one dataset to another.
Specifically, the measurement modeling framework can help us examine how factors related to how coreference is defined and annotated might affect the measurement of the constructs of \textit{coreference}, \textit{CR model performance}, or \textit{CR model generalization}.

\section{Coreference}
\label{sec:datasets}

\textit{Coreference} is itself an unobservable construct; therefore, measurements of coreference must be inferred from indicators, e.g., annotations of coreference in a dataset.

In this section, we first describe the coreference datasets we consider in our analysis (\S\ref{sec:intro-datasets}). We then describe discrepancies in how coreference is defined (\S\ref{sec:constructs}) and operationalized (\S\ref{sec:annotation-guidelines}) within these datasets.

\subsection{Datasets}
\label{sec:intro-datasets}

Our goal is to understand whether multi-dataset evaluations tell us anything about models' ability to generalize.
To examine this, we focus our analysis on datasets that have been used in prior work to evaluate CR model generalization. Specifically, our selection of datasets is based on prior multi-dataset evaluations of CR models \citep{toshniwal-etal-2021-generalization,xia-van-durme-2021-moving,zhu-etal-2021-ontogum,zabokrtsky-etal-2023-findings}. In total, we consider seven English-language datasets containing annotations of identity coreference at a document level (see Table~\ref{tab:dataset-stats} for dataset figures). Appendix \ref{sec:scope} provides more details regarding the selection of these datasets, how they have been preprocessed or formatted, and the use of these datasets within the broader coreference literature.

\para{OntoNotes} OntoNotes 5.0~\citep{ontonotes5} consists of news, conversations, web data, and biblical text in which coreference was annotated by experts. We use the English CoNLL-2012 Shared Task version of this dataset~\citep{pradhan-etal-2012-conll}. While OntoNotes 5.0 is annotated for multiple phenomena, the CoNLL-2012 Shared Task contains only so-called ``identical'' coreference, also sometimes referred to as ``anaphoric coreference.''

\begin{table}[t]
    \centering
    \scriptsize
    \def\arraystretch{0.65}
    \setlength{\tabcolsep}{1.1em}
    \begin{tabular}{lcccc}
        Dataset & Train & Dev. & Test & Total Words (K) \\
        \midrule
        OntoNotes & 2,802 & 343 & 348 & 1,632 \\[.2em]
        PreCo & 36,120 & 500 & 500 & 12,493 \\[.2em]
        Phrase Det. & 695 & 45 & 45 & 1,321 \\[.2em]
        \hdashline\\[-.5em]
        OntoGUM & 165 & 24 & 24 & 204 \\[.2em]
        LitBank & 80 & 10 & 10 & 211 \\[.2em]
        ARRAU & 444 & 33 & 75 & 348 \\[.2em]
        MMC & 955 & 134 & 133 & 324 \\
        \bottomrule
    \end{tabular}
    \caption{Number of documents for each dataset split.}
    \label{tab:dataset-stats}
    \vspace{-10pt}
\end{table}

\para{PreCo} PreCo \citep{chen-etal-2018-preco} consists of English comprehensive exams annotated for coreference by trained university students.

\para{Phrase Detectives} Phrase Detectives 3.0~\citep{yu-etal-2023-aggregating} consists of Wikipedia, fiction, and technical text. The training set annotations were sourced by aggregating annotations of users playing the \textit{Phrase Detectives} online game, where the users were tasked with annotating coreferences, or verifying others' annotations. The test set was annotated by experts. We use the ``CoNLL'' formatted version of the dataset.\footnote{\url{https://github.com/dali-ambiguity/Phrase-Detectives-Corpus-3.0}}

\para{OntoGUM} OntoGUM~\citep{zhu-etal-2021-ontogum} is a reformatted version of the GUM corpus~\citep{10.1007/s10579-016-9343-x}. The GUM corpus was originally annotated in an iterative process by linguistics students. OntoGUM was created by transforming GUM, using deterministic rules and annotated syntactic parses, to follow the OntoNotes annotation guidelines. We use version 9.2.0 of OntoGUM.

\para{LitBank} LitBank~\citep{ bamman-etal-2020-annotated} consists of coreference annotated in English literature by experts. Only noun phrases of ACE categories and pronouns have been annotated for coreference.\footnote{The ACE categories are: people, facilities, geo-political entities, locations, vehicles, and organizations.} We use the ``CoNLL'' formatted dataset version.\footnote{\url{https://github.com/dbamman/litbank}}

\para{ARRAU} ARRAU 2.1~\citep{Uryupina_Artstein_Bristot_Cavicchio_Delogu_Rodriguez_Poesio_2020} is a dataset of written news and spoken conversations annotated for various anaphoric phenomenon by experts.
We use all documents and the formatting procedure of \citet{xia-van-durme-2021-moving} which keeps only the coarsest-grained  annotations.

\paragraph{MMC} Multilingual Coreference Resolution in Multiparty Dialogue (MMC)~\citep{zheng-etal-2023-multilingual} is a dataset of television transcripts. The training set was annotated by crowdworkers, and the test set by experts.
We use the English portion of the ``CoNLL'' formatted version of this dataset.\footnote{\url{https://github.com/boyuanzheng010/mmc}}

\begin{table*}[ht]
    \centering
    \scriptsize
    \def\arraystretch{0.65}
    \setlength{\tabcolsep}{0.01em}
    \begin{tabular}{p{0.13\linewidth}@{\hskip 1em} p{0.3\linewidth}@{\hskip 1em} p{0.25\linewidth}@{\hskip 1em} p{0.25\linewidth}}
        Type & OntoNotes & PreCo & Phrase Detectives \\
        \midrule
        \textbf{Generic Mentions} \newline [dogs] can bark & Generics mentions are only annotated when they corefer with a pronoun or determinate noun phrase, or when they occur in a news headline. & All generic noun phrases and modifiers can be annotated as coreferring. & All generic noun phrases can be annotated as coreferring. \\
        \midrule
        \textbf{Verb Phrases} \newline it will [grow] & The head of a verb phrase can be annotated as coreferring with a determinant noun phrase. & Not annotated. & Not annotated. \\
        \midrule
        \textbf{Appositives} \newline [[Abe] , [the chef]] & Annotated in the dataset, but not considered coreference. & Annotated as three mentions: both noun phrases and the larger span. & Annotated in the dataset, but not considered coreference. \\
        \midrule
        \textbf{Copular Predicates} \newline [he] is [the teacher] & Not annotated. & In a copular structure, the referent and attribute are annotated as coreferring. & Annotated in the dataset, but not considered coreference. \\
        \midrule
        \midrule
        \textbf{Nesting} \newline [he [himself]] & When two nested mentions share a head, only the dominant mention is annotated. Proper nouns cannot contain nested mentions. & Appositives and mentions with shared heads are annotated as nested mentions. & The right-most mention in an appositive is considered referring to a distinct entity and can therefore be annotated as a nested mention. No restrictions on nesting of proper nouns. \\
        \midrule
        \textbf{Compound \newline Modifiers} \newline [Taiwan] authorities & Compound modifiers are annotated if non-adjective proper nouns that are not a nationality acronym. & All compound modifiers can be annotated as coreferring. & No explicit restrictions on the annotation of compound modifiers. \\
        \bottomrule
    \end{tabular}
    \caption{Noted differences in how coreference is defined and operationalized in the training datasets.}
    \label{tab:general-cr-differences}
\end{table*}

\subsection{Differences Across Datasets}
Here, we overview differences in how coreference is defined and operationalized across  the three largest datasets---OntoNotes, PreCo, and Phrase Detectives---which we also use as training sets.

\subsubsection{Differing Definitions}
\label{sec:constructs}

Because datasets differ in how coreference is conceptualized, measurements of how CR models generalize from one dataset to another might be confounded by or rather capture the 
differences in the corresponding definitions of coreference.

Notable differences in how these three datasets conceptualize coreference primarily stem from whether the following phenomena are considered to be coreference relations: 
\textbf{1)} multiple \textbf{generic} expressions that could possibly be interpreted as referring to the same discourse entity (i.e., generic only), 
\textbf{2)} expressions that could be interpreted as referring to the same event where at least one is a \textbf{verb phrase} (VPs), 
\textbf{3)} two expressions in \textbf{apposition}, and 
\textbf{4)} two expressions in a \textbf{copular structure}.
We overview whether datasets include these in the definition of coreference in Table~\ref{tab:general-cr-differences}.

These differences in how coreference is defined were determined based on the original documentation of the respective datasets. We considered both the original publication as well as the annotation guidelines where available. More details on the definitions of coreference for each dataset are provided in Appendix~\ref{sec:detailed-constructs}.

\subsubsection{Differing Annotations}
\label{sec:annotation-guidelines}

In addition to differences in how coreference is conceptualized, we also consider aspects related to how the construct is operationalized. 

While there are many dimensions to how coreference is operationalized within each dataset, we focus on differences in datasets' annotation guidelines, particularly concerning how coreferring expressions are specified to be annotated. We highlight two key ways in which annotation differs between the three training datasets which can be empirically studied: \textbf{1)} the annotation of \textbf{nested mentions} and \textbf{2)} the annotation of \textbf{compound modifiers}. We provide more details of the differences between datasets in Table~\ref{tab:general-cr-differences}. 

The definition of coreference influences the operationalization, and therefore differences in annotations and definitions are closely coupled. For example, the inclusion of apposition in the definition of coreference will affect how nested mentions are then annotated.

In fact, in our analysis we often consider factors related to the operationalization and conceptualization of coreference jointly as these are often not explicitly delineated in the datasets' documentations. 
More generally, there is little or no clear distinction in the literature between decisions related to conceptualizing coreference---that is, defining the theoretical construct being measured---versus decisions related to the measurement of coreference---how a given definition is being operationalized.

For instance, certain decisions are explicitly made to increase accuracy in terms of annotator agreement; in the case of annotating verb phrases in OntoNotes ``[o]nly the single-word head of the verb phrase is included in the span, even in cases where the entire verb phrase is the logical co-referent''~\citep{bbn2007guidelines}. However, dimensions such as the handling of compound modifiers could be argued to be either a way to clarify how annotators should approach their task and thus improve annotator accuracy, 
or part of the definition of coreference itself. Clarifying this distinction is a community-wide research endeavor.

\section{Methodology}
\label{sec:experiments}

This section first provides a high-level overview of the experimental setup (\S\ref{sec:disaggregated-eval}), and then describes the details of how the model evaluation is performed (\S\ref{sec:eval-metrics}), as well as the models we consider (\S\ref{sec:models}). 
Our goal is to understand how measurements of generalization might be affected by differing definitions and operationalizations of coreference across datasets. To this end, we empirically examine performance variations for types of coreference that differ in how they are conceptualized and/or operationalized across datasets.

\subsection{Disaggregated Evaluation}
\label{sec:disaggregated-eval}

The purpose of disaggregated evaluations is to understand whether or not failures in generalization, as measured by a model's accuracy on multiple test sets, are correlated with differences in how CR is defined and operationalized in these test sets. 
For instance, if measurements of generalization seem indeed sensitive to differences in how coreference is defined across datasets, this might indicate 
that measurements of generalization capture or are confounded by
differences in how coreference is defined rather than solely capturing the capacity of a CR model to generalize some consistent conceptualization of coreference.\looseness=-1

We evaluate models on the test set of the dataset they were trained on, which we refer to as \textit{in-domain}, as well as the test sets of all other datasets, which we refer to as \textit{out-of-domain}. We then highlight types of coreference that appear significantly more difficult for a model to infer out-of-domain (different definitions or operationalizations) as compared to in-domain (same definition and operationalization). These cases are typically interpreted as indicating limited generalization, and we 
we want to examine whether such limited generalization is correlated with types of coreference that differ in the ways in which they are conceptualized and operationalized.

More formally, let $\theta$ be the parameterization of some model, $X$ the in-domain dataset, $\hat{X}$ some out-of-domain dataset, and $f_\theta(X)$ some measurement of accuracy on dataset $X$. The \textit{aggregate generalization gap} (AGG) is defined to be:
\begin{equation}
    \text{AGG} = \lvert f_\theta(X) - f_\theta(\hat{X}) \rvert 
\end{equation}

Further, let $X_t$ be the subset of dataset $X$ that is of type $t$ for which we can calculate model accuracy as $f_\theta(X_t)$. We will highlight cases where the aggregate generalization gap is significantly different from the \textit{type generalization gap} (TGG) for a given type of coreference $t$, defined to be:
\begin{equation}
    \text{TGG} = \lvert f_\theta(X_t) - f_\theta(\hat{X_t}) \rvert
\end{equation}

\subsubsection{Formalizing Coreference Types}

Here, we clarify the precise differences in how coreference is defined and operationalized that we consider in our study. We formalize these differences as particular types of coreference that vary in how they are handled between datasets. All of the differences in coreference annotation that we have noted are with respect to the annotation of particular types of mentions that occur in a coreference relation (Table~\ref{tab:general-cr-differences}), therefore our definition for each \textit{type of coreference} is based on first formally defining particular \textit{types of mentions}. For now, it is not important whether a given mention is annotated in the dataset or predicted by a model, as we will clarify which is the case when performing evaluations of both precision and recall (\S\ref{sec:eval-metrics}).

\para{Metadata}
In practice, we identify some mention types using universal dependency relations \citep{nivre-etal-2020-universal} or part-of-speech (POS) tags. This metadata is sourced as follows: for OntoNotes, we convert the expert annotated constituency parses into a dependency parse using the Stanford CoreNLP library \citep{manning-etal-2014-stanford}. For OntoGUM, we use the existing expert annotated dependency parses. For PreCo, Phrase Detectives, and all other datasets, we parse the dataset using the Stanza parser \citep{qi2020stanza}.

\vspace{8pt}

\para{Studied Coreference Types}

\begin{enumerate}[wide, labelindent=0pt]
    \item \textbf{Nested}: {\em A mention that corefers with an overlapping span (either larger or smaller).} \vspace{-2pt}
    \begin{quote}
        \small
        President Chen said, [he [himself]] has not returned to his hometown \ \ldots
    \end{quote} \vspace{-4pt}

    \item \textbf{ON Generic}: {\em A coreferring mention that is a generic noun phrase.} We use the OntoNotes definition of generic, which we refer to as an ON generic: a noun phrase is considered generic in OntoNotes if it has an indefinite article or is plural with no article. We check if there exists \textit{a} or \textit{an} as the child of a \texttt{det} relation, or there is no \texttt{det} relation and the POS tag is \texttt{NNS} (plural noun).
    \vspace{-2pt}
    \begin{quote}
        \small
        \ldots you know yeah they had [a farm] when they were first married \ldots I don't know how many years they had [a dairy farm] \ldots
    \end{quote} \vspace{-2pt}
    
    \item \textbf{Compound}: {\em A coreferring mention that is a compound modifier.} Defined to be a mention that is the dependent of a \texttt{compound:nn} relation.
    \vspace{-2pt}
        \begin{quote}
        \small
        \ldots \ we miss our [Taiwan] compatriots even more, and \ldots \ gave a speech, expressing hopes that [Taiwan] authorities would \ \ldots
    \end{quote}  \vspace{-2pt}
    
    \item \textbf{Copular}: {\em A mention that is in a copular construction with another mention to which it corefers.} Two mentions are said to be in a copular construction if they are in an \texttt{nsubj} relation with each other and the rightmost mention is the head of a \texttt{cop} dependency relation.
    \vspace{-2pt}
    \begin{quote}
        \small
        Yet I realize that in my father's eyes, [I] will always be [his little girl].
    \end{quote} \vspace{-2pt}
\end{enumerate}

\subsubsection{Evaluation Metrics}
\label{sec:eval-metrics}

In this section, we outline the specific metrics used to calculate the accuracy measurement $f_\theta(\cdot)$. To evaluate a model, we would like to know how well said model's predicted coreference clusters agree with the datasets' annotated clusters.

\para{Notation} Let document $D = (w_1, w_2, \ldots, w_n)$ be a sequence of $n$ words. Let $S = \{ s_{i:j} \mid 1 \leq i \leq j \leq n \}$ denote the set of all spans of words in $D$, where a span is a continuous subsequence of words $s_{i:j} = (w_i, \ldots, w_j)$.
The set of annotated mentions $M \subseteq S$ denotes all spans in $S$ that are annotated as coreferring with at least one other span. The set of entities $E = \{ E_1, E_2, \ldots, E_k \}$ is a $k$-partition of $M$, i.e., a family of $k$ non-empty, pairwise disjoint subsets of $S$. Each entity $E_i \in E$ represents a set of spans that refer to the same discourse entity in the document. A CR model takes as input $D$ and outputs $E' = \{ E'_1, \ldots, E'_\ell \}$, a prediction of the true $E$, which is an $\ell$-partition of the set of model predicted mentions $M'$.

\para{Metrics} There is not a single standard approach for calculating disaggregated CR evaluation scores. We choose to the adopt the B$^3$-based method of \citet{bamman-etal-2020-annotated} due to its clear intuition, although in practice we find that our conclusions are not sensitive to the exact choice of metric.

In the aggregate case, we report agreement using B$^3$ recall, precision, and F1 \citep{bagga-baldwin-1998-entity-based}.
The probabilistic intuition behind B$^3$ recall is that it calculates, in expectation over all annotated mentions, the probability that an annotated coreferring mention is correctly recalled by the model. A similar intuition applies to precision.

More formally, B$^3$ calculates cluster agreement by averaging over all mentions in the cluster. Given mention $s$, let $\mathcal{E} = \bigcup \{E_i \mid s \in E_i\}$ be the set of all mentions annotated as coreferring with $s$, and $\mathcal{E'} = \bigcup \{E'_i \mid s \in E'_i\}$ the set of all mentions predicted as coreferring with $s$. B$^3$ calculates recall (R) and precision (P) as

{\small
\begin{equation}
    \text{R} = \frac{1}{|M|} \sum_{s \in M} \frac{|\mathcal{E} \cap \mathcal{E'}|}{|\mathcal{E}|} \qquad \text{P} = \frac{1}{|M'|} \sum_{s \in M'} \frac{|\mathcal{E'} \cap \mathcal{E}|}{|\mathcal{E'}|}
\end{equation}
}
from which F1 is calculated as the harmonic mean.

\begin{table*}[t]
    \centering
    \scriptsize
    \def\arraystretch{0.65}
    \begin{tabular}{lccccccccccccr}
    \multirow{3}{*}{\backslashbox{Test}{Train \vspace{2em}}} & \multicolumn{6}{c}{ON} & \multicolumn{3}{c}{PC} & \multicolumn{3}{c}{PD} & \\
    \cmidrule(lr){2-7}\cmidrule(lr){8-10}\cmidrule(lr){11-13}
& \multicolumn{3}{c}{LinkAppend} & \multicolumn{3}{c}{LingMess\textsubscript{ON}} & \multicolumn{3}{c}{LingMess\textsubscript{PC}} & \multicolumn{3}{c}{LingMess\textsubscript{PD}} \\
 \cmidrule(lr){2-4} \cmidrule(lr){5-7} \cmidrule(lr){8-10} \cmidrule(lr){11-13} & R & P & F1 & R & P & F1 & R & P & F1 & R & P & F1 & \quad \# Ments.  \\
\midrule
OntoNotes (CoNLL) & 83.7 & 82.9 & 83.3 & 79.2 & 82.6 & 80.9 & 64.1 & 48.8 & 55.4 & 33.5 & 61.4 & 43.3 & \\[.2em]
OntoNotes (B$^3$) & 83.2 & 82.0 & 82.6 & 77.6 & 81.6 & 79.5 & 62.5 & 45.8 & 52.9 & 28.0 & 57.5 & 37.7 & 19,764 \\[.2em]
\hdashline
\rule{0pt}{8pt}\quad Nested & 43.9 & 70.6 & 54.1 & 48.5 & 65.2 & 55.6 & 20.7 & 2.0 & \ccell{3.6} & \hphantom00.1 & \hphantom00.2 & \ccell{0.2} & 247 \\[.2em]
\quad ON Generic & 56.1 & 64.7 & 60.1 & 47.0 & 63.8 & 54.1 & 52.9 & 8.1 & \ccell{14.0} & 34.0 & 13.8 & 19.7 & 1,182 \\[.2em]
\quad Compound & 81.3 & 76.1 & 78.6 & 81.2 & 80.9 & 81.0 & 73.5 & 14.3 & 23.9 & \hphantom00.0 & \hphantom00.0 & \ccell{0.0} & 412 \\[.2em]
\quad Copular & 83.3 & 45.5 & 58.8 & 66.7 & 40.0 & 50.0 & 40.5 & \hphantom00.4 & \ccell{0.8} & \hphantom00.0 & \hphantom00.0 & 0.0 & 12 \\[.2em]
\midrule
OntoGUM (CoNLL) & 64.6 & 78.3 & 70.7 & 59.3 & 76.3 & 66.7 & 57.3 & 49.3 & 52.9 & 38.5 & 67.2 & 48.9 &  \\[.2em]
OntoGUM (B$^3$) & 65.0 & 75.6 & 69.9 & 59.3 & 73.6 & 65.7 & 58.3 & 46.0 & 51.4 & 33.0 & 61.8 & 43.0 & 2,707 \\[.2em]
\hdashline
\rule{0pt}{8pt}\quad Nested & 25.0 & 2.1 & 3.8 & \hphantom00.0 & \hphantom00.0 & \hphantom00.0 & \hphantom00.0 & \hphantom00.0 & \hphantom00.0 & \hphantom00.0 & \hphantom00.0 & \hphantom00.0 & 2 \\[.2em]
\quad ON Generic & 34.5 & 50.0 & 40.8 & 18.4 & 23.4 & 20.4 & 64.4 & \hphantom03.9 & \hphantom07.3 & 21.0 & \hphantom03.5 & \hphantom06.0 & 169 \\[.2em]
\quad Compound & 50.6 & 69.2 & 58.1 & 35.4 & 74.3 & 47.5 & 48.7 & 13.5 & 21.2 & \hphantom00.0 & \hphantom00.0 & \hphantom00.0 & 64 \\[.2em]
\bottomrule
    \end{tabular}
    \vspace{-8pt}
    \caption{
    Aggregated and disaggregated metrics intended to measure model performance. Scores are correlated with differences in how coreference is defined and operationalized across datasets, indicating that purported measurements of out-of-domain generalization also encompass these differences between datasets. Each column corresponds to a model trained on the respective training set: OntoNotes (ON), PreCo (PC), or Phrase Detectives (PD). For LinkAppend~\citep{bohnet-etal-2023-coreference}, we use the publicly released weights. For LingMess~\citep{otmazgin2022lingmess}, we train the model on each of the three training sets. Each row corresponds to the specified type of coreference. F1 scores are highlighted for types of coreference where the F1 score dropped significantly out-of-domain as compared to the overall difference in aggregate scores. “\# Ments.” refers to the number of annotated mentions.
    \label{tab:results-ontonotes}
    }
    \vspace{-8pt}
\end{table*}

\begin{table}[ht]
    \centering
    \scriptsize
    \def\arraystretch{0.65}
    \begin{tabular}{lcccr}
 & \multicolumn{3}{c}{LingMess} & \\
 \cmidrule(lr){2-4}
 Eval. & \multicolumn{1}{c}{ON} & \multicolumn{1}{c}{PC} & \multicolumn{1}{c}{PD} & \# Ments.  \\
\midrule
PreCo (C) & 64.5 & 77.1 & 50.8 & \\[.2em]
PreCo (B$^3$) & 64.2 & 76.3 & 47.3 & 25,983 \\[.2em]
\hdashline
\rule{0pt}{8pt}\quad Nested & \ccell{11.6} & 55.9 & \ccell{0.5} & 1,760 \\[.2em]
\quad ON Generic & \ccell{12.0} & 70.5 & 50.9 & 3,520 \\[.2em]
\quad Compound & \ccell{5.0} & 77.1 & \ccell{0.0} & 2,163 \\[.2em]
\quad Copular & \ccell{0.8} & 63.7 & \ccell{0.0} & 1,410 \\
\midrule
Phrase Det. (C) & 70.8 & 60.9 & 44.3 & \\[.2em]
Phrase Det. (B$^3$) & 69.8 & 56.9 & 36.8 & 3,259 \\[.2em]
\hdashline
\rule{0pt}{8pt}\quad Nested & 27.9 & 3.4 & 4.0 & 74 \\[.2em]
\quad ON Generic & \ccell{8.3} & 42.9 & 44.4 & 296 \\[.2em]
\quad Compound & 8.1 & 26.6 & \hphantom00.0 & 127 \\[.2em]
\quad Copular & \hphantom00.0 & \hphantom00.7 & \hphantom00.0 & 10 \\
\midrule
LitBank (C) & 65.5 & 50.9 & 36.6 & \\[.2em]
LitBank (B$^3$) & 66.3 & 47.0 & 32.9 & 2,267 \\[.2em]
\hdashline
\rule{0pt}{8pt}\quad Nested & 24.3 & 7.3 & \hphantom00.0 & 39 \\[.2em]
\quad ON Generic & 5.7 & 16.5 & 13.2 & 91 \\[.2em]
\quad Compound & \hphantom00.0 & \hphantom00.0 & \hphantom00.0 & 11 \\[.2em]
\quad Copular & 28.6 & 6.7 & \hphantom00.0 & 10 \\
\midrule
ARRAU (C) & 59.4 & 53.5 & 38.1 & \\[.2em]
ARRAU (B$^3$) & 57.8 & 51.2 & 33.2 & 6,132 \\[.2em]
\hdashline
\rule{0pt}{8pt}\quad Nested & 7.0 & 1.3 & 1.1 & 284 \\[.2em]
\quad ON Generic & 7.3 & 48.4 & 40.2 & 924 \\[.2em]
\quad Compound & 36.2 & 55.3 & \hphantom00.0 & 545 \\[.2em]
\quad Copular & 13.1 & 2.3 & 13.9 & 32 \\
\midrule
MMC (C) & 61.4 & 55.6 & 45.2 & \\[.2em]
MMC (B$^3$) & 57.3 & 51.0 & 40.5 & 5,338 \\[.2em]
\hdashline
\rule{0pt}{8pt}\quad Nested & 10.0 & 11.0 & \hphantom00.0 & 140 \\[.2em]
\quad ON Generic & 16.6 & 53.0 & 34.8 & 393 \\[.2em]
\quad Compound & \hphantom00.0 & 10.3 & 100.0 & 7 \\[.2em]
\quad Copular & 1.2 & 55.1 & 1.2 & 636 \\
\bottomrule
    \end{tabular}
    \vspace{-8pt}
    \caption{Disaggregated F1 scores using the same setup as Table~\ref{tab:results-ontonotes}, as evaluated on additional test sets. F1 scores are highlighted when the difference relative to the in-domain model is significantly larger than the same difference calculated for all mentions. C indicates CoNLL F1 score.}
    \label{tab:results-all}
    \vspace{-8pt}
\end{table}

To then evaluate performance on a given type of coreference, we use a modified B$^3$ score that calculates the expected agreement only for mentions of the given type. That is, we only sum over mentions $s$ that are of the type of mention being considered. In other words, $M$ is replaced with $\{m \mid m \in M \land \text{is\_type}(m)\}$ when calculating recall, and $M'$ is replaced with $\{m \mid m \in M' \land \text{is\_type}(m)\}$ when calculating precision. Where here, $\text{is\_type}(m)$ indicates that mention $m$ is of the type being considered. We also report  CoNLL score \citep{denis2009global,pradhan-etal-2012-conll} in the aggregate case to facilitate comparison with prior work.

\subsection{Models}
\label{sec:models}

We focus our evaluation on four models: the publicly available LinkAppend model~\citep{bohnet-etal-2023-coreference} and three versions of LingMess~\citep{otmazgin-etal-2023-lingmess},  each trained on one of the training datasets. These are currently the best performing decoder and encoder-based models on the OntoNotes test set. For all models, hyperparameters are presented in Appendix \ref{sec:model-hps}.

\para{LinkAppend} \citet{bohnet-etal-2023-coreference} proposed a method for formulating CR modeling as a sequence-to-sequence task where the objective is to predict links between mentions---forming a new coreference cluster---or to append mentions to existing clusters. For this, they trained a 13B parameter mT5 language model \citep{xue-etal-2021-mt5}, with speaker and genre metadata, which achieved state-of-the-art performance on OntoNotes. We use the publicly released weights of the best performing trained model which we refer to as LinkAppend.

\para{LingMess} \citet{otmazgin2022lingmess} proposed an encoder-based CR model architecture and training procedure based on the mention ranking paradigm. They used this training recipe to train a state-of-the-art model, named LingMess, on the OntoNotes dataset. This model was initialized with a Longformer-large model as the pretrained encoder \citep{Beltagy2020Longformer}.
To evaluate LingMess, we train a model on each of the three training sets using the default hyperparameters and Longformer-large encoder. All training runs take at most 24 hours on a single 40GB A100 GPU.

\section{Results}
\label{sec:results}

To examine how model performance varies across different types of coreference and different datasets, we conduct an extensive disaggregated evaluation.
Summary results are presented in Table~\ref{tab:results-ontonotes} for models evaluated on OntoNotes and OntoGUM, and Table~\ref{tab:results-all} for all other test sets (exact AGG and TGG values are presented in Appendix~\ref{sec:addt-results}).

We illustrate how model failures correlate with differences between training and test sets in the way coreference is both defined and operationalized.
In some of the datasets we studied (e.g., PreCo), the coreference types we consider in our study are fairly prevalent and represent up to 34\% of all mentions annotated as coreferring (Table~\ref{tab:results-all}).

\para{We observe significant performance variations for types of coreference that are defined differently across datasets} 
Consider the coreference types highlighted in Table~\ref{tab:results-ontonotes}. 
There are, for instance, significant performance variations (e.g., a performance drop of over 40 percentage points) for both generic and copular mentions between
the in-domain---i.e., trained on a dataset constructed using the same definition of coreference as the test dataset---and out-of-domain---i.e., trained on datasets constructed using a different definition of coreference---LingMess models. 
While this variation might be explained by a number of factors, both PreCo and Phrase Detectives (used to train the out-of-domain models) differ from OntoNotes in that these types of mentions are considered part of the definition of coreference in both of these datasets.
As a result, for instance, while only a few copular structures are annotated as coreferring in OntoNotes, LingMess\textsubscript{PC} predicts 1,158 coreferring mentions in a copular structure; \textit{e.g.,} this example of a copular structure correctly predicted (according to the definitions used by PreCo and Phrase Detectives) as coreferring by LingMess\textsubscript{PC}:
\vspace{-2pt}
\begin{quote}
    \small
    \ldots [One of the two honorable guests in the studio]\textsubscript{1} is [Professor Zhou Hanhug]\textsubscript{1} \ldots
\end{quote} \vspace{-4pt}
 
Consequently, both estimations of performance as well as measurements of model generalization will necessarily depend on which definition is used.

\para{We also observe performance variations when there are operationalization differences across datasets constructed using the same definition of coreference} 
Consider the case of OntoGUM in Table~\ref{tab:results-ontonotes}, a dataset constructed using the same definition of coreference as OntoNotes. 
This allows us to examine the impact of potential differences in how the same definition of coreference was operationalized across datasets. 
Notably, models trained on OntoNotes have poor performance on OntoGUM nested mentions despite these datasets also using the same annotation guidelines.
While it is hard to fully disentangle why these differences exists, we conjecture they might be in part due to how these guidelines were interpreted by annotators: for instance, the OntoGUM paper describes removing i-within-i coreference relations; upon manual inspection of the cases where a model predicts a nested mention in the OntoGUM dataset that is not in the OntoGUM annotations, many of these purported failures appear to be attributable to this removal as all models predict certain i-within-i constructions.
Consider the following nested coreference relations correctly predicted by LinkAppend, but not annotated in OntoGUM:
\vspace{-2pt}
\begin{quote}
    \small
    Open-air markets, bookstores, and a [a Bart station all [its]\textsubscript{1} own]\textsubscript{1} make Rockridge another of many pleasant stops \ldots \\[1em]
    So there's nothing too special about [the bed [itself]\textsubscript{1}]\textsubscript{1}, but it does have a very important function \ldots 
\end{quote} \vspace{-2pt}

Such differences in how the same type of coreference are operationalized by dataset builders can further distort measurements of both model performance and model generalization.

\section{Discussion}

One might find it intuitive that measurements of CR model performance and generalization are likely to be impacted by differences in how coreference is defined and operationalized across datasets; and yet, existing multi-dataset evaluations do not explicitly considered how these differences might impact the performance estimates and the conclusions we can draw about models ability to generalize.  

Existing evaluations of model generalization across datasets mostly report accuracy using aggregate metrics \citep{moosavi-strube-2018-using,bamman-etal-2020-annotated,toshniwal-etal-2021-generalization}. Such aggregations, however, can obscure certain systematic errors \citep{10.1145/3531146.3533233}. Understanding of what exactly is being measured requires more explicit consideration of the measurement process. For this reason, we believe measurement modeling provides a useful framework for analyzing measurements in CR.

Our findings also highlight how relying on aggregated evaluations can obscure why models fail to generalize across datasets. %
We find this to be especially true in the case of coreference which involves multiple sub-phenomena that are inconsistently annotated across datasets. If we do not resolve inconsistencies and errors in existing datasets, it will remain difficult to accurately interpret performance evaluation results based on these datasets.

\subsection{Relative Ranking of Model Accuracy}

A question one might ask is whether we can still make valid conclusions regarding the relative ranking of models' accuracy despite the absolute scores suffering from the issues with measurement validity that we have shown. Ultimately, however, ranking models based on empirical indicators is still a form of measurement and will therefore be subject to the same considerations that we have raised; \textit{e.g.,} Is there a consistent definition of the theoretical concept being measured? Are the measurements valid in that they encompass only those variables we are intending to measure?

Consider the following empirical example, comparing LinkAppend and LingMess\textsubscript{ON}: on 4 of the 5 “out-of-domain” test sets from Table~\ref{tab:results-all}, LinkAppend has a higher CoNLL F1 score than LingMess\textsubscript{ON}. On the LitBank test set, however, LingMess\textsubscript{ON} has a 65.5 CoNLL F1 score versus 64.3 for LinkAppend. A possible explanation for this relative ranking is that LingMess\textsubscript{ON} has a lower recall across datasets (e.g., predicts fewer mentions in general), and LitBank is only annotated for a restricted set of mention types. Therefore, underfitting on the OntoNotes training set might lead to a better score on LitBank due to differences in how coreference is defined in these datasets rather than the ability of a model to generalize to unseen coreference links. Comparing the relative ranking of models will thus have limited "convergent validity" (\textit{do measurements correlate with other measurements of the same construct?}) \citep{jacobs2021measurement} because the ranking is not consistent across datasets.\looseness=-1

\section{Conclusion}
\label{sec:conclusion}

We propose viewing the measurement of coreference, CR model performance, and CR model generalization from the perspective of measurement modeling. To do so requires clearly distinguishing the theoretical construct intended to be measured from its measurement via a measurement model. This allows one to more clearly consider what is being measured, and possible concerns with the validity of these measurements.

Through a disaggregated, out-of-domain evaluation of CR models, we have shown that models appear systematically limited in their ability to generalize to certain types of coreference that differ in how they are defined or operationalized across datasets. We showed that models ostensibly display limited generalization for types of coreference even when evaluated on datasets intended to use the same annotation guidelines.

If the research goal is to develop CR models that generalize a consistent conceptualization across datasets, iteratively improving in-domain performance will have limited impact, as models' ability to generalize will inherently depend on operationalization differences between datasets and evaluation schemes.\footnote{Our code is available at \url{https://github.com/ianporada/coref-data}}\looseness=-1

\para{Measurement Modeling as a Blueprint} Finally, our work provides a blueprint for evaluating out-of-domain performance in a way that accounts for known inconsistencies in how coreference is defined and operationalized across datasets. In particular, our work highlights how explicitly considering the definition and operationalization of the construct being measured, as well as disaggregating performance results, can be used to better understand what we can learn from multi-dataset evaluations about models' ability to generalize. This is a necessary step in meaningfully evaluating models of contested constructs across multiple datasets.

\subsection{Limitations and Future Work}
\label{sec:limitations}
Our empirical study only establishes correlations between model performance and types of CR when assessed across and within datasets.
While there are certainly many more types of coreference that could be considered, by focusing on a select few types, we are able to study their interaction with dataset operationalization in greater detail.

Future work could explore other dimensions of operationalization such as: 1) the data processing (e.g., tokenization method); 2) the CR task format; 3) and the annotation procedure, including noise and ambiguity in annotation. Ultimately, a more holistic understanding of the way that coreference is operationalized can provide better insights into model accuracy and guide future modeling decisions. A promising direction might be to define such differences using grounded theory \citep{strauss1997grounded} similar to the method used by \citet{robertson-et-al-chi-2021}.

Furthermore, while one might intuitively expect that simply taking the intersection of all coreference types annotated in existing corpora can be used to measure model accuracy, the inclusion of certain phenomenon is not always discrete or easily determinable. In addition, taking an intersection of existing annotations does not solve of the problem of needing to clearly define the construct being measured. Rather, focusing on datasets with minimal differences is a possible direction for untangling the impact of any individual factor.

\section{Ethics Statement}
\label{sec:ethics-statement}

We have focused on evaluation datasets of CR, but we did not quantify possible downstream implications of these findings. Different types of CR might also be more prevalent in certain types of corpora that might be about or written by minoritized groups, or that might cover sensitive topics.  
CR models have been shown to exhibit biases, inferring coreferences disparately for distinct social groups \citep{webster-etal-2018-mind,Kocijan_Camburu_Lukasiewicz_2021,hossain-etal-2023-misgendered}. Similarly, CR and other NLP datasets are also known to contain biased, stereotypical, or in other way problematic context~\cite{cao2021toward,selvam2022tail}.

\section*{Acknowledgements}

The authors acknowledge the material support of NVIDIA in the form of computational resources. Ian Porada is supported by a fellowship from the Fonds de recherche du Québec (FRQ).

\bibliography{anthology,custom}

\begin{thebibliography}{84}
\expandafter\ifx\csname natexlab\endcsname\relax\def\natexlab#1{#1}\fi

\bibitem[{Adcock and Collier(2001)}]{adcock_collier_2001}
Robert Adcock and David Collier. 2001.
\newblock \href {https://doi.org/10.1017/S0003055401003100} {Measurement validity: A shared standard for qualitative and quantitative research}.
\newblock \emph{American Political Science Review}, 95(3):529–546.

\bibitem[{Bagga and Baldwin(1998)}]{bagga-baldwin-1998-entity-based}
Amit Bagga and Breck Baldwin. 1998.
\newblock \href {https://doi.org/10.3115/980845.980859} {Entity-based cross-document coreferencing using the vector space model}.
\newblock In \emph{36th Annual Meeting of the Association for Computational Linguistics and 17th International Conference on Computational Linguistics, Volume 1}, pages 79--85, Montreal, Quebec, Canada. Association for Computational Linguistics.

\bibitem[{Bamman et~al.(2020)Bamman, Lewke, and Mansoor}]{bamman-etal-2020-annotated}
David Bamman, Olivia Lewke, and Anya Mansoor. 2020.
\newblock \href {https://aclanthology.org/2020.lrec-1.6} {An annotated dataset of coreference in {E}nglish literature}.
\newblock In \emph{Proceedings of the Twelfth Language Resources and Evaluation Conference}, pages 44--54, Marseille, France. European Language Resources Association.

\bibitem[{{BBN Technologies}(2007)}]{bbn2007guidelines}
{BBN Technologies}. 2007.
\newblock \href {https://www.ldc.upenn.edu/sites/www.ldc.upenn.edu/files/english-coreference-guidelines.pdf} {Co-reference guidelines for english ontonotes. {Version 7.0.}}
\newblock Guidelines accompanying OntoNotes 5.0 data release.

\bibitem[{Beltagy et~al.(2020)Beltagy, Peters, and Cohan}]{Beltagy2020Longformer}
Iz~Beltagy, Matthew~E. Peters, and Arman Cohan. 2020.
\newblock Longformer: The long-document transformer.
\newblock \emph{arXiv:2004.05150}.

\bibitem[{Bhattacherjee(2012)}]{bhattacherjee2012social}
Anol Bhattacherjee. 2012.
\newblock \emph{Social science research: Principles, methods, and practices}.
\newblock USA.

\bibitem[{Black(1999)}]{black_doing_1999}
Thomas~R. Black. 1999.
\newblock Doing quantitative research in the social sciences: {An} integrated approach to research design, measurement and statistics.
\newblock \emph{Doing quantitative research in the social sciences: An integrated approach to research design, measurement and statistics.}, pages xiv, 751--xiv, 751.
\newblock Place: Thousand Oaks, CA Publisher: Sage Publications Ltd.

\bibitem[{Blodgett et~al.(2021)Blodgett, Lopez, Olteanu, Sim, and Wallach}]{blodgett-etal-2021-stereotyping}
Su~Lin Blodgett, Gilsinia Lopez, Alexandra Olteanu, Robert Sim, and Hanna Wallach. 2021.
\newblock \href {https://doi.org/10.18653/v1/2021.acl-long.81} {Stereotyping {N}orwegian salmon: An inventory of pitfalls in fairness benchmark datasets}.
\newblock In \emph{Proceedings of the 59th Annual Meeting of the Association for Computational Linguistics and the 11th International Joint Conference on Natural Language Processing (Volume 1: Long Papers)}, pages 1004--1015, Online. Association for Computational Linguistics.

\bibitem[{Bohnet et~al.(2023)Bohnet, Alberti, and Collins}]{bohnet-etal-2023-coreference}
Bernd Bohnet, Chris Alberti, and Michael Collins. 2023.
\newblock \href {https://doi.org/10.1162/tacl_a_00543} {Coreference resolution through a seq2seq transition-based system}.
\newblock \emph{Transactions of the Association for Computational Linguistics}, 11:212--226.

\bibitem[{Bollen(2001)}]{smelser_indicator_2001}
K.~A. Bollen. 2001.
\newblock \href {https://doi.org/https://doi.org/10.1016/B0-08-043076-7/00709-9} {Measurement models}.
\newblock In Neil~J. Smelser and Paul~B. Baltes, editors, \emph{Indicator: {Methodology}}, International {Encyclopedia} of the {Social} \& {Behavioral} {Sciences}, chapter~2, pages 7282--7287. Pergamon, Oxford.

\bibitem[{Cao and Daum{\'e}~III(2021)}]{cao2021toward}
Yang~Trista Cao and Hal Daum{\'e}~III. 2021.
\newblock Toward gender-inclusive coreference resolution: An analysis of gender and bias throughout the machine learning lifecycle.
\newblock \emph{Computational Linguistics}, 47(3):615--661.

\bibitem[{Chai et~al.(2022)Chai, Moosavi, Gurevych, and Strube}]{chai-etal-2022-evaluating}
Haixia Chai, Nafise~Sadat Moosavi, Iryna Gurevych, and Michael Strube. 2022.
\newblock \href {https://aclanthology.org/2022.crac-1.7} {Evaluating coreference resolvers on community-based question answering: From rule-based to state of the art}.
\newblock In \emph{Proceedings of the Fifth Workshop on Computational Models of Reference, Anaphora and Coreference}, pages 61--73, Gyeongju, Republic of Korea. Association for Computational Linguistics.

\bibitem[{Chai and Strube(2023)}]{chai-strube-2023-investigating}
Haixia Chai and Michael Strube. 2023.
\newblock \href {https://doi.org/10.18653/v1/2023.findings-emnlp.671} {Investigating multilingual coreference resolution by universal annotations}.
\newblock In \emph{Findings of the Association for Computational Linguistics: EMNLP 2023}, pages 10010--10024, Singapore. Association for Computational Linguistics.

\bibitem[{Chen et~al.(2018)Chen, Fan, Lu, Yuille, and Rong}]{chen-etal-2018-preco}
Hong Chen, Zhenhua Fan, Hao Lu, Alan Yuille, and Shu Rong. 2018.
\newblock \href {https://doi.org/10.18653/v1/D18-1016} {{P}re{C}o: A large-scale dataset in preschool vocabulary for coreference resolution}.
\newblock In \emph{Proceedings of the 2018 Conference on Empirical Methods in Natural Language Processing}, pages 172--181, Brussels, Belgium. Association for Computational Linguistics.

\bibitem[{Chinchor(1992)}]{chinchor-1992-statistical}
Nancy Chinchor. 1992.
\newblock \href {https://aclanthology.org/M92-1003} {The statistical significance of the {MUC}-4 results}.
\newblock In \emph{{F}ourth {M}essage {U}nderstanding {C}onference ({MUC}-4): Proceedings of a Conference Held in {M}c{L}ean, {V}irginia, {J}une 16-18, 1992}.

\bibitem[{Clark(1977)}]{clark1977bridging}
H.~H. Clark. 1977.
\newblock \emph{Bridging}, pages 411--420. Cambridge University Press.

\bibitem[{Cybulska and Vossen(2014)}]{cybulska-vossen-2014-using}
Agata Cybulska and Piek Vossen. 2014.
\newblock \href {http://www.lrec-conf.org/proceedings/lrec2014/pdf/840_Paper.pdf} {Using a sledgehammer to crack a nut? lexical diversity and event coreference resolution}.
\newblock In \emph{Proceedings of the Ninth International Conference on Language Resources and Evaluation ({LREC}'14)}, pages 4545--4552, Reykjavik, Iceland. European Language Resources Association (ELRA).

\bibitem[{Dasigi et~al.(2019)Dasigi, Liu, Marasovi{\'c}, Smith, and Gardner}]{dasigi-etal-2019-quoref}
Pradeep Dasigi, Nelson~F. Liu, Ana Marasovi{\'c}, Noah~A. Smith, and Matt Gardner. 2019.
\newblock \href {https://doi.org/10.18653/v1/D19-1606} {{Q}uoref: A reading comprehension dataset with questions requiring coreferential reasoning}.
\newblock In \emph{Proceedings of the 2019 Conference on Empirical Methods in Natural Language Processing and the 9th International Joint Conference on Natural Language Processing (EMNLP-IJCNLP)}, pages 5925--5932, Hong Kong, China. Association for Computational Linguistics.

\bibitem[{Denis and Baldridge(2009)}]{denis2009global}
Pascal Denis and Jason Baldridge. 2009.
\newblock Global joint models for coreference resolution and named entity classification.
\newblock \emph{Procesamiento del lenguaje natural}, 42.

\bibitem[{Doddington et~al.(2004)Doddington, Mitchell, Przybocki, Ramshaw, Strassel, and Weischedel}]{doddington-etal-2004-automatic}
George Doddington, Alexis Mitchell, Mark Przybocki, Lance Ramshaw, Stephanie Strassel, and Ralph Weischedel. 2004.
\newblock \href {http://www.lrec-conf.org/proceedings/lrec2004/pdf/5.pdf} {The automatic content extraction ({ACE}) program {--} tasks, data, and evaluation}.
\newblock In \emph{Proceedings of the Fourth International Conference on Language Resources and Evaluation ({LREC}{'}04)}, Lisbon, Portugal. European Language Resources Association (ELRA).

\bibitem[{Durrett and Klein(2013)}]{durrett-klein-2013-easy}
Greg Durrett and Dan Klein. 2013.
\newblock \href {https://aclanthology.org/D13-1203} {Easy victories and uphill battles in coreference resolution}.
\newblock In \emph{Proceedings of the 2013 Conference on Empirical Methods in Natural Language Processing}, pages 1971--1982, Seattle, Washington, USA. Association for Computational Linguistics.

\bibitem[{Edes(1968)}]{edes1968output}
Ellen Edes. 1968.
\newblock Output conditions on anaphoric expressions with split antecedents.
\newblock \emph{unpublished paper, Harvard University}.

\bibitem[{Emami et~al.(2019)Emami, Trichelair, Trischler, Suleman, Schulz, and Cheung}]{emami-etal-2019-knowref}
Ali Emami, Paul Trichelair, Adam Trischler, Kaheer Suleman, Hannes Schulz, and Jackie Chi~Kit Cheung. 2019.
\newblock \href {https://doi.org/10.18653/v1/P19-1386} {The {K}now{R}ef coreference corpus: Removing gender and number cues for difficult pronominal anaphora resolution}.
\newblock In \emph{Proceedings of the 57th Annual Meeting of the Association for Computational Linguistics}, pages 3952--3961, Florence, Italy. Association for Computational Linguistics.

\bibitem[{Gandhi et~al.(2022)Gandhi, Field, and Strubell}]{https://doi.org/10.48550/arxiv.2210.07602}
Nupoor Gandhi, Anjalie Field, and Emma Strubell. 2022.
\newblock \href {https://doi.org/10.48550/ARXIV.2210.07602} {Mention annotations alone enable efficient domain adaptation for coreference resolution}.

\bibitem[{Grishman and Sundheim(1996)}]{grishman-sundheim-1996-design}
Ralph Grishman and Beth Sundheim. 1996.
\newblock \href {https://doi.org/10.3115/1119018.1119072} {Design of the {MUC}-6 evaluation}.
\newblock In \emph{TIPSTER TEXT PROGRAM PHASE II: Proceedings of a Workshop held at Vienna, Virginia, May 6-8, 1996}, pages 413--422, Vienna, Virginia, USA. Association for Computational Linguistics.

\bibitem[{Hirst(1981)}]{hirst1981anaphora}
Graeme Hirst. 1981.
\newblock \href {https://doi.org/10.1007/3-540-10858-0_2} {\emph{Anaphora}}, pages 4--32. Springer Berlin Heidelberg, Berlin, Heidelberg.

\bibitem[{Hobbs(1978)}]{hobbs1978resolving}
Jerry~R Hobbs. 1978.
\newblock Resolving pronoun references.
\newblock \emph{Lingua}, 44(4):311--338.

\bibitem[{Hossain et~al.(2023)Hossain, Dev, and Singh}]{hossain-etal-2023-misgendered}
Tamanna Hossain, Sunipa Dev, and Sameer Singh. 2023.
\newblock \href {https://doi.org/10.18653/v1/2023.acl-long.293} {{MISGENDERED}: Limits of large language models in understanding pronouns}.
\newblock In \emph{Proceedings of the 61st Annual Meeting of the Association for Computational Linguistics (Volume 1: Long Papers)}, pages 5352--5367, Toronto, Canada. Association for Computational Linguistics.

\bibitem[{Hutchinson et~al.(2022)Hutchinson, Rostamzadeh, Greer, Heller, and Prabhakaran}]{10.1145/3531146.3533233}
Ben Hutchinson, Negar Rostamzadeh, Christina Greer, Katherine Heller, and Vinodkumar Prabhakaran. 2022.
\newblock \href {https://doi.org/10.1145/3531146.3533233} {Evaluation gaps in machine learning practice}.
\newblock In \emph{Proceedings of the 2022 ACM Conference on Fairness, Accountability, and Transparency}, FAccT '22, page 1859–1876, New York, NY, USA. Association for Computing Machinery.

\bibitem[{Jacobs and Wallach(2021)}]{jacobs2021measurement}
Abigail~Z. Jacobs and Hanna Wallach. 2021.
\newblock \href {https://doi.org/10.1145/3442188.3445901} {Measurement and fairness}.
\newblock In \emph{Proceedings of the 2021 ACM Conference on Fairness, Accountability, and Transparency}, FAccT '21, page 375–385, New York, NY, USA. Association for Computing Machinery.

\bibitem[{Kantor(1977)}]{kantor1977management}
Robert~Neal Kantor. 1977.
\newblock \emph{The management and comprehension of discourse connection by pronouns in English.}
\newblock The Ohio State University.

\bibitem[{Kline(2011)}]{kline_principles_2011}
R.B. Kline. 2011.
\newblock \href {https://books.google.ca/books?id=mGf3Ex59AX0C} {\emph{Principles and {Practice} of {Structural} {Equation} {Modeling}}}.
\newblock Methodology in the social sciences. Guilford Publications.

\bibitem[{Kocijan et~al.(2021)Kocijan, Camburu, and Lukasiewicz}]{Kocijan_Camburu_Lukasiewicz_2021}
Vid Kocijan, Oana-Maria Camburu, and Thomas Lukasiewicz. 2021.
\newblock \href {https://doi.org/10.1609/aaai.v35i14.17557} {The gap on gap: Tackling the problem of differing data distributions in bias-measuring datasets}.
\newblock \emph{Proceedings of the AAAI Conference on Artificial Intelligence}, 35(14):13180--13188.

\bibitem[{K{\"u}bler and Zhekova(2011)}]{kubler-zhekova-2011-singletons}
Sandra K{\"u}bler and Desislava Zhekova. 2011.
\newblock \href {https://aclanthology.org/R11-1036} {Singletons and coreference resolution evaluation}.
\newblock In \emph{Proceedings of the International Conference Recent Advances in Natural Language Processing 2011}, pages 261--267, Hissar, Bulgaria. Association for Computational Linguistics.

\bibitem[{Lapshinova-Koltunski et~al.(2022)Lapshinova-Koltunski, Ferreira, Lartaud, and Hardmeier}]{lapshinova-koltunski-etal-2022-parcorfull2}
Ekaterina Lapshinova-Koltunski, Pedro~Augusto Ferreira, Elina Lartaud, and Christian Hardmeier. 2022.
\newblock \href {https://aclanthology.org/2022.lrec-1.85} {{P}ar{C}or{F}ull2.0: a parallel corpus annotated with full coreference}.
\newblock In \emph{Proceedings of the Thirteenth Language Resources and Evaluation Conference}, pages 805--813, Marseille, France. European Language Resources Association.

\bibitem[{Lu and Ng(2018)}]{lu2018event}
Jing Lu and Vincent Ng. 2018.
\newblock Event coreference resolution: A survey of two decades of research.
\newblock In \emph{IJCAI}, pages 5479--5486.

\bibitem[{Lu and Ng(2020)}]{lu-ng-2020-conundrums}
Jing Lu and Vincent Ng. 2020.
\newblock \href {https://doi.org/10.18653/v1/2020.emnlp-main.536} {Conundrums in entity coreference resolution: Making sense of the state of the art}.
\newblock In \emph{Proceedings of the 2020 Conference on Empirical Methods in Natural Language Processing (EMNLP)}, pages 6620--6631, Online. Association for Computational Linguistics.

\bibitem[{Manning et~al.(2014)Manning, Surdeanu, Bauer, Finkel, Bethard, and McClosky}]{manning-etal-2014-stanford}
Christopher Manning, Mihai Surdeanu, John Bauer, Jenny Finkel, Steven Bethard, and David McClosky. 2014.
\newblock \href {https://doi.org/10.3115/v1/P14-5010} {The {S}tanford {C}ore{NLP} natural language processing toolkit}.
\newblock In \emph{Proceedings of 52nd Annual Meeting of the Association for Computational Linguistics: System Demonstrations}, pages 55--60, Baltimore, Maryland. Association for Computational Linguistics.

\bibitem[{Moosavi et~al.(2019)Moosavi, Born, Poesio, and Strube}]{moosavi-etal-2019-using}
Nafise~Sadat Moosavi, Leo Born, Massimo Poesio, and Michael Strube. 2019.
\newblock \href {https://doi.org/10.18653/v1/P19-1408} {Using automatically extracted minimum spans to disentangle coreference evaluation from boundary detection}.
\newblock In \emph{Proceedings of the 57th Annual Meeting of the Association for Computational Linguistics}, pages 4168--4178, Florence, Italy. Association for Computational Linguistics.

\bibitem[{Moosavi and Strube(2017)}]{moosavi-strube-2017-lexical}
Nafise~Sadat Moosavi and Michael Strube. 2017.
\newblock \href {https://doi.org/10.18653/v1/P17-2003} {Lexical features in coreference resolution: To be used with caution}.
\newblock In \emph{Proceedings of the 55th Annual Meeting of the Association for Computational Linguistics (Volume 2: Short Papers)}, pages 14--19, Vancouver, Canada. Association for Computational Linguistics.

\bibitem[{Moosavi and Strube(2018)}]{moosavi-strube-2018-using}
Nafise~Sadat Moosavi and Michael Strube. 2018.
\newblock \href {https://doi.org/10.18653/v1/D18-1018} {Using linguistic features to improve the generalization capability of neural coreference resolvers}.
\newblock In \emph{Proceedings of the 2018 Conference on Empirical Methods in Natural Language Processing}, pages 193--203, Brussels, Belgium. Association for Computational Linguistics.

\bibitem[{Nedoluzhko et~al.(2021)Nedoluzhko, Nov{\'a}k, Popel, {\v{Z}}abokrtsk{\`y}, and Zeman}]{nedoluzhko2021coreference}
Anna Nedoluzhko, Michal Nov{\'a}k, Martin Popel, Zdenek {\v{Z}}abokrtsk{\`y}, and Daniel Zeman. 2021.
\newblock Coreference meets universal dependencies--a pilot experiment on harmonizing coreference datasets for 11 languages.
\newblock \emph{{\'U}FAL MFF UK, Praha, Czechia}.

\bibitem[{Nivre et~al.(2020)Nivre, de~Marneffe, Ginter, Haji{\v{c}}, Manning, Pyysalo, Schuster, Tyers, and Zeman}]{nivre-etal-2020-universal}
Joakim Nivre, Marie-Catherine de~Marneffe, Filip Ginter, Jan Haji{\v{c}}, Christopher~D. Manning, Sampo Pyysalo, Sebastian Schuster, Francis Tyers, and Daniel Zeman. 2020.
\newblock \href {https://aclanthology.org/2020.lrec-1.497} {{U}niversal {D}ependencies v2: An evergrowing multilingual treebank collection}.
\newblock In \emph{Proceedings of the Twelfth Language Resources and Evaluation Conference}, pages 4034--4043, Marseille, France. European Language Resources Association.

\bibitem[{Otmazgin et~al.(2022)Otmazgin, Cattan, and Goldberg}]{otmazgin2022lingmess}
Shon Otmazgin, Arie Cattan, and Yoav Goldberg. 2022.
\newblock \href {http://arxiv.org/abs/2205.12644} {Lingmess: Linguistically informed multi expert scorers for coreference resolution}.

\bibitem[{Otmazgin et~al.(2023)Otmazgin, Cattan, and Goldberg}]{otmazgin-etal-2023-lingmess}
Shon Otmazgin, Arie Cattan, and Yoav Goldberg. 2023.
\newblock \href {https://doi.org/10.18653/v1/2023.eacl-main.202} {{L}ing{M}ess: Linguistically informed multi expert scorers for coreference resolution}.
\newblock In \emph{Proceedings of the 17th Conference of the European Chapter of the Association for Computational Linguistics}, pages 2752--2760, Dubrovnik, Croatia. Association for Computational Linguistics.

\bibitem[{Peng et~al.(2015)Peng, Khashabi, and Roth}]{peng-etal-2015-solving}
Haoruo Peng, Daniel Khashabi, and Dan Roth. 2015.
\newblock \href {https://doi.org/10.3115/v1/N15-1082} {Solving hard coreference problems}.
\newblock In \emph{Proceedings of the 2015 Conference of the North {A}merican Chapter of the Association for Computational Linguistics: Human Language Technologies}, pages 809--819, Denver, Colorado. Association for Computational Linguistics.

\bibitem[{Poesio et~al.(1999)Poesio, Bruneseaux, and Romary}]{poesio-etal-1999-mate}
M.~Poesio, F.~Bruneseaux, and L.~Romary. 1999.
\newblock \href {https://aclanthology.org/W99-0309} {The {MATE} meta-scheme for coreference in dialogues in multiple languages}.
\newblock In \emph{Towards Standards and Tools for Discourse Tagging}.

\bibitem[{Poesio et~al.(2023)Poesio, Yu, Paun, Aloraini, Lu, Haber, and Cokal}]{poesio2023computational}
Massimo Poesio, Juntao Yu, Silviu Paun, Abdulrahman Aloraini, Pengcheng Lu, Janosch Haber, and Derya Cokal. 2023.
\newblock Computational models of anaphora.
\newblock \emph{Annual Review of Linguistics}, 9:561--587.

\bibitem[{Poot and van Cranenburgh(2020)}]{poot-van-cranenburgh-2020-benchmark}
Corb{\`e}n Poot and Andreas van Cranenburgh. 2020.
\newblock \href {https://aclanthology.org/2020.crac-1.9} {A benchmark of rule-based and neural coreference resolution in {D}utch novels and news}.
\newblock In \emph{Proceedings of the Third Workshop on Computational Models of Reference, Anaphora and Coreference}, pages 79--90, Barcelona, Spain (online). Association for Computational Linguistics.

\bibitem[{Pradhan et~al.(2012)Pradhan, Moschitti, Xue, Uryupina, and Zhang}]{pradhan-etal-2012-conll}
Sameer Pradhan, Alessandro Moschitti, Nianwen Xue, Olga Uryupina, and Yuchen Zhang. 2012.
\newblock \href {https://aclanthology.org/W12-4501} {{C}o{NLL}-2012 shared task: Modeling multilingual unrestricted coreference in {O}nto{N}otes}.
\newblock In \emph{Joint Conference on {EMNLP} and {C}o{NLL} - Shared Task}, pages 1--40, Jeju Island, Korea. Association for Computational Linguistics.

\bibitem[{Qi et~al.(2020)Qi, Zhang, Zhang, Bolton, and Manning}]{qi2020stanza}
Peng Qi, Yuhao Zhang, Yuhui Zhang, Jason Bolton, and Christopher~D. Manning. 2020.
\newblock \href {https://nlp.stanford.edu/pubs/qi2020stanza.pdf} {Stanza: A {Python} natural language processing toolkit for many human languages}.
\newblock In \emph{Proceedings of the 58th Annual Meeting of the Association for Computational Linguistics: System Demonstrations}.

\bibitem[{Rahman and Ng(2012)}]{rahman-ng-2012-resolving}
Altaf Rahman and Vincent Ng. 2012.
\newblock \href {https://aclanthology.org/D12-1071} {Resolving complex cases of definite pronouns: The {W}inograd schema challenge}.
\newblock In \emph{Proceedings of the 2012 Joint Conference on Empirical Methods in Natural Language Processing and Computational Natural Language Learning}, pages 777--789, Jeju Island, Korea. Association for Computational Linguistics.

\bibitem[{Robertson et~al.(2021)Robertson, Olteanu, Diaz, Shokouhi, and Bailey}]{robertson-et-al-chi-2021}
Ronald~E Robertson, Alexandra Olteanu, Fernando Diaz, Milad Shokouhi, and Peter Bailey. 2021.
\newblock \href {https://doi.org/10.1145/3411764.3445557} {“i can’t reply with that”: Characterizing problematic email reply suggestions}.
\newblock In \emph{Proceedings of the 2021 CHI Conference on Human Factors in Computing Systems}, CHI '21, New York, NY, USA. Association for Computing Machinery.

\bibitem[{Roesiger et~al.(2018)Roesiger, Riester, and Kuhn}]{roesiger-etal-2018-bridging}
Ina Roesiger, Arndt Riester, and Jonas Kuhn. 2018.
\newblock \href {https://aclanthology.org/C18-1298} {Bridging resolution: Task definition, corpus resources and rule-based experiments}.
\newblock In \emph{Proceedings of the 27th International Conference on Computational Linguistics}, pages 3516--3528, Santa Fe, New Mexico, USA. Association for Computational Linguistics.

\bibitem[{Selvam et~al.(2022)Selvam, Dev, Khashabi, Khot, and Chang}]{selvam2022tail}
Nikil~Roashan Selvam, Sunipa Dev, Daniel Khashabi, Tushar Khot, and Kai-Wei Chang. 2022.
\newblock The tail wagging the dog: Dataset construction biases of social bias benchmarks.
\newblock \emph{arXiv preprint arXiv:2210.10040}.

\bibitem[{Straka(2023)}]{straka-2023-ufal}
Milan Straka. 2023.
\newblock \href {https://doi.org/10.18653/v1/2023.crac-sharedtask.4} {{{\'U}FAL} {C}or{P}ipe at {CRAC} 2023: Larger context improves multilingual coreference resolution}.
\newblock In \emph{Proceedings of the CRAC 2023 Shared Task on Multilingual Coreference Resolution}, pages 41--51, Singapore. Association for Computational Linguistics.

\bibitem[{Strauss and Corbin(1997)}]{strauss1997grounded}
Anselm Strauss and Juliet~M Corbin. 1997.
\newblock \emph{Grounded theory in practice}.
\newblock Sage.

\bibitem[{Subramanian and Roth(2019)}]{subramanian-roth-2019-improving}
Sanjay Subramanian and Dan Roth. 2019.
\newblock \href {https://doi.org/10.18653/v1/S19-1021} {Improving generalization in coreference resolution via adversarial training}.
\newblock In \emph{Proceedings of the Eighth Joint Conference on Lexical and Computational Semantics (*{SEM} 2019)}, pages 192--197, Minneapolis, Minnesota. Association for Computational Linguistics.

\bibitem[{Sukthanker et~al.(2020)Sukthanker, Poria, Cambria, and Thirunavukarasu}]{SUKTHANKER2020139}
Rhea Sukthanker, Soujanya Poria, Erik Cambria, and Ramkumar Thirunavukarasu. 2020.
\newblock \href {https://doi.org/https://doi.org/10.1016/j.inffus.2020.01.010} {Anaphora and coreference resolution: A review}.
\newblock \emph{Information Fusion}, 59:139--162.

\bibitem[{Toshniwal et~al.(2021)Toshniwal, Xia, Wiseman, Livescu, and Gimpel}]{toshniwal-etal-2021-generalization}
Shubham Toshniwal, Patrick Xia, Sam Wiseman, Karen Livescu, and Kevin Gimpel. 2021.
\newblock \href {https://doi.org/10.18653/v1/2021.crac-1.12} {On generalization in coreference resolution}.
\newblock In \emph{Proceedings of the Fourth Workshop on Computational Models of Reference, Anaphora and Coreference}, pages 111--120, Punta Cana, Dominican Republic. Association for Computational Linguistics.

\bibitem[{Urbizu et~al.(2019)Urbizu, Soraluze, and Arregi}]{urbizu-etal-2019-deep}
Gorka Urbizu, Ander Soraluze, and Olatz Arregi. 2019.
\newblock \href {https://doi.org/10.18653/v1/W19-2806} {Deep cross-lingual coreference resolution for less-resourced languages: The case of {B}asque}.
\newblock In \emph{Proceedings of the Second Workshop on Computational Models of Reference, Anaphora and Coreference}, pages 35--41, Minneapolis, USA. Association for Computational Linguistics.

\bibitem[{Uryupina(2008)}]{uryupina-2008-error}
Olga Uryupina. 2008.
\newblock \href {http://www.lrec-conf.org/proceedings/lrec2008/pdf/487_paper.pdf} {Error analysis for learning-based coreference resolution}.
\newblock In \emph{Proceedings of the Sixth International Conference on Language Resources and Evaluation ({LREC}'08)}, Marrakech, Morocco. European Language Resources Association (ELRA).

\bibitem[{Uryupina et~al.(2020)Uryupina, Artstein, Bristot, Cavicchio, Delogu, Rodriguez, and Poesio}]{Uryupina_Artstein_Bristot_Cavicchio_Delogu_Rodriguez_Poesio_2020}
Olga Uryupina, Ron Artstein, Antonella Bristot, Federica Cavicchio, Francesca Delogu, Kepa~J. Rodriguez, and Massimo Poesio. 2020.
\newblock \href {https://doi.org/10.1017/S1351324919000056} {Annotating a broad range of anaphoric phenomena, in a variety of genres: the arrau corpus}.
\newblock \emph{Natural Language Engineering}, 26(1):95–128.

\bibitem[{Versley(2008)}]{Versley2008}
Yannick Versley. 2008.
\newblock \href {https://doi.org/10.1007/s11168-008-9059-1} {Vagueness and referential ambiguity in a large-scale annotated corpus}.
\newblock \emph{Research on Language and Computation}, 6(3):333--353.

\bibitem[{Webber(1991)}]{webber1991structure}
Bonnie~Lynn Webber. 1991.
\newblock Structure and ostension in the interpretation of discourse deixis.
\newblock \emph{Language and Cognitive processes}, 6(2):107--135.

\bibitem[{Webster et~al.(2018)Webster, Recasens, Axelrod, and Baldridge}]{webster-etal-2018-mind}
Kellie Webster, Marta Recasens, Vera Axelrod, and Jason Baldridge. 2018.
\newblock \href {https://doi.org/10.1162/tacl_a_00240} {Mind the {GAP}: A balanced corpus of gendered ambiguous pronouns}.
\newblock \emph{Transactions of the Association for Computational Linguistics}, 6:605--617.

\bibitem[{Weischedel et~al.(2013)}]{ontonotes5}
Ralph Weischedel et~al. 2013.
\newblock Ontonotes release 5.0.

\bibitem[{Winograd(1972)}]{winograd1972understanding}
Terry Winograd. 1972.
\newblock Understanding natural language.
\newblock \emph{Cognitive psychology}, 3(1):1--191.

\bibitem[{Xia and Van~Durme(2021)}]{xia-van-durme-2021-moving}
Patrick Xia and Benjamin Van~Durme. 2021.
\newblock \href {https://doi.org/10.18653/v1/2021.emnlp-main.425} {Moving on from {O}nto{N}otes: Coreference resolution model transfer}.
\newblock In \emph{Proceedings of the 2021 Conference on Empirical Methods in Natural Language Processing}, pages 5241--5256, Online and Punta Cana, Dominican Republic. Association for Computational Linguistics.

\bibitem[{Xue et~al.(2021)Xue, Constant, Roberts, Kale, Al-Rfou, Siddhant, Barua, and Raffel}]{xue-etal-2021-mt5}
Linting Xue, Noah Constant, Adam Roberts, Mihir Kale, Rami Al-Rfou, Aditya Siddhant, Aditya Barua, and Colin Raffel. 2021.
\newblock \href {https://doi.org/10.18653/v1/2021.naacl-main.41} {m{T}5: A massively multilingual pre-trained text-to-text transformer}.
\newblock In \emph{Proceedings of the 2021 Conference of the North American Chapter of the Association for Computational Linguistics: Human Language Technologies}, pages 483--498, Online. Association for Computational Linguistics.

\bibitem[{Yang et~al.(2012)Yang, Mao, Xiang, Tsang, Chai, and Chieu}]{yang-etal-2012-domain}
Jian~Bo Yang, Qi~Mao, Qiao~Liang Xiang, Ivor Wai-Hung Tsang, Kian Ming~Adam Chai, and Hai~Leong Chieu. 2012.
\newblock \href {https://aclanthology.org/D12-1068} {Domain adaptation for coreference resolution: An adaptive ensemble approach}.
\newblock In \emph{Proceedings of the 2012 Joint Conference on Empirical Methods in Natural Language Processing and Computational Natural Language Learning}, pages 744--753, Jeju Island, Korea. Association for Computational Linguistics.

\bibitem[{Yu et~al.(2023{\natexlab{a}})Yu, Nov{\'a}k, Aloraini, Moosavi, Paun, Pradhan, and Poesio}]{yu-etal-2023-universal}
Juntao Yu, Michal Nov{\'a}k, Abdulrahman Aloraini, Nafise~Sadat Moosavi, Silviu Paun, Sameer Pradhan, and Massimo Poesio. 2023{\natexlab{a}}.
\newblock \href {https://aclanthology.org/2023.iwcs-1.19} {The universal anaphora scorer 2.0}.
\newblock In \emph{Proceedings of the 15th International Conference on Computational Semantics}, pages 183--194, Nancy, France. Association for Computational Linguistics.

\bibitem[{Yu et~al.(2023{\natexlab{b}})Yu, Paun, Camilleri, Garcia, Chamberlain, Kruschwitz, and Poesio}]{yu-etal-2023-aggregating}
Juntao Yu, Silviu Paun, Maris Camilleri, Paloma Garcia, Jon Chamberlain, Udo Kruschwitz, and Massimo Poesio. 2023{\natexlab{b}}.
\newblock \href {https://doi.org/10.18653/v1/2023.eacl-main.54} {Aggregating crowdsourced and automatic judgments to scale up a corpus of anaphoric reference for fiction and {W}ikipedia texts}.
\newblock In \emph{Proceedings of the 17th Conference of the European Chapter of the Association for Computational Linguistics}, pages 767--781, Dubrovnik, Croatia. Association for Computational Linguistics.

\bibitem[{Yuan et~al.(2022)Yuan, Xia, May, Van~Durme, and Boyd-Graber}]{yuan-etal-2022-adapting}
Michelle Yuan, Patrick Xia, Chandler May, Benjamin Van~Durme, and Jordan Boyd-Graber. 2022.
\newblock \href {https://doi.org/10.18653/v1/2022.acl-long.519} {Adapting coreference resolution models through active learning}.
\newblock In \emph{Proceedings of the 60th Annual Meeting of the Association for Computational Linguistics (Volume 1: Long Papers)}, pages 7533--7549, Dublin, Ireland. Association for Computational Linguistics.

\bibitem[{{\v{Z}}abokrtsk{\'y} et~al.(2023){\v{Z}}abokrtsk{\'y}, Konopik, Nedoluzhko, Nov{\'a}k, Ogrodniczuk, Popel, Prazak, Sido, and Zeman}]{zabokrtsky-etal-2023-findings}
Zden{\v{e}}k {\v{Z}}abokrtsk{\'y}, Miloslav Konopik, Anna Nedoluzhko, Michal Nov{\'a}k, Maciej Ogrodniczuk, Martin Popel, Ondrej Prazak, Jakub Sido, and Daniel Zeman. 2023.
\newblock \href {https://doi.org/10.18653/v1/2023.crac-sharedtask.1} {Findings of the second shared task on multilingual coreference resolution}.
\newblock In \emph{Proceedings of the CRAC 2023 Shared Task on Multilingual Coreference Resolution}, pages 1--18, Singapore. Association for Computational Linguistics.

\bibitem[{{\v{Z}}abokrtsk{\'y} et~al.(2022){\v{Z}}abokrtsk{\'y}, Konop{\'\i}k, Nedoluzhko, Nov{\'a}k, Ogrodniczuk, Popel, Pra{\v{z}}{\'a}k, Sido, Zeman, and Zhu}]{zabokrtsky-etal-2022-findings}
Zden{\v{e}}k {\v{Z}}abokrtsk{\'y}, Miloslav Konop{\'\i}k, Anna Nedoluzhko, Michal Nov{\'a}k, Maciej Ogrodniczuk, Martin Popel, Ond{\v{r}}ej Pra{\v{z}}{\'a}k, Jakub Sido, Daniel Zeman, and Yilun Zhu. 2022.
\newblock \href {https://aclanthology.org/2022.crac-mcr.1} {Findings of the shared task on multilingual coreference resolution}.
\newblock In \emph{Proceedings of the CRAC 2022 Shared Task on Multilingual Coreference Resolution}, pages 1--17, Gyeongju, Republic of Korea. Association for Computational Linguistics.

\bibitem[{Zeldes(2017)}]{10.1007/s10579-016-9343-x}
Amir Zeldes. 2017.
\newblock \href {https://doi.org/10.1007/s10579-016-9343-x} {The gum corpus: Creating multilayer resources in the classroom}.
\newblock \emph{Lang. Resour. Eval.}, 51(3):581–612.

\bibitem[{Zeldes(2022)}]{can_we_fix}
Amir Zeldes. 2022.
\newblock \href {https://doi.org/10.5210/dad.2022.102} {Opinion piece: Can we fix the scope for coreference?}
\newblock \emph{Dialogue \& Discourse}, 13(1):41–62.

\bibitem[{Zeldes and Zhang(2016)}]{zeldes-zhang-2016-annotation}
Amir Zeldes and Shuo Zhang. 2016.
\newblock \href {https://doi.org/10.18653/v1/W16-0713} {When annotation schemes change rules help: A configurable approach to coreference resolution beyond {O}nto{N}otes}.
\newblock In \emph{Proceedings of the Workshop on Coreference Resolution Beyond {O}nto{N}otes ({CORBON} 2016)}, pages 92--101, San Diego, California. Association for Computational Linguistics.

\bibitem[{Zhao and Ng(2014)}]{zhao-ng-2014-domain}
Shanheng Zhao and Hwee~Tou Ng. 2014.
\newblock \href {https://doi.org/10.3115/v1/W14-1104} {Domain adaptation with active learning for coreference resolution}.
\newblock In \emph{Proceedings of the 5th International Workshop on Health Text Mining and Information Analysis (Louhi)}, pages 21--29, Gothenburg, Sweden. Association for Computational Linguistics.

\bibitem[{Zheng et~al.(2023)Zheng, Xia, Yarmohammadi, and Van~Durme}]{zheng-etal-2023-multilingual}
Boyuan Zheng, Patrick Xia, Mahsa Yarmohammadi, and Benjamin Van~Durme. 2023.
\newblock \href {https://doi.org/10.1162/tacl_a_00581} {Multilingual coreference resolution in multiparty dialogue}.
\newblock \emph{Transactions of the Association for Computational Linguistics}, 11:922--940.

\bibitem[{Zhou and Choi(2018)}]{zhou-choi-2018-exist}
Ethan Zhou and Jinho~D. Choi. 2018.
\newblock \href {https://aclanthology.org/C18-1003} {They exist! introducing plural mentions to coreference resolution and entity linking}.
\newblock In \emph{Proceedings of the 27th International Conference on Computational Linguistics}, pages 24--34, Santa Fe, New Mexico, USA. Association for Computational Linguistics.

\bibitem[{Zhu et~al.(2021)Zhu, Pradhan, and Zeldes}]{zhu-etal-2021-ontogum}
Yilun Zhu, Sameer Pradhan, and Amir Zeldes. 2021.
\newblock \href {https://doi.org/10.18653/v1/2021.acl-short.59} {{O}nto{GUM}: Evaluating contextualized {SOTA} coreference resolution on 12 more genres}.
\newblock In \emph{Proceedings of the 59th Annual Meeting of the Association for Computational Linguistics and the 11th International Joint Conference on Natural Language Processing (Volume 2: Short Papers)}, pages 461--467, Online. Association for Computational Linguistics.

\bibitem[{Zinsmeister and Dipper(2010)}]{Zinsmeister2010TowardsAS}
Heike Zinsmeister and Stefanie Dipper. 2010.
\newblock \href {https://api.semanticscholar.org/CorpusID:13075189} {Towards a standard for annotating abstract anaphora}.
\newblock In \emph{Workshop on Language Resource and Language Technology Standards at the International Conference on Language Resources and Evaluation}.

\end{thebibliography}

\appendix

\section{Additional Dataset Details}

\subsection{Dataset Selection and Scope}
\label{sec:scope}

\subsubsection{Dataset Selection}

Our selection of datasets is based on the multi-dataset evaluations of \citet{toshniwal-etal-2021-generalization}, \citet{xia-van-durme-2021-moving}, \citet{zhu-etal-2021-ontogum}, and \citet{zabokrtsky-etal-2023-findings}.

Of the datasets evaluated by \citet{xia-van-durme-2021-moving} (\xia), we select the same large-scale training sets, OntoNotes and PreCo, as well as the two largest English-language test sets, LitBank and ARRAU. The collection of \citet{toshniwal-etal-2021-generalization} is mostly a subset of \xia, with the exception of the Friends dataset~\citep{zhou-choi-2018-exist}. We use the more recent MMC dataset~\citep{zheng-etal-2023-multilingual} which encompasses the Friends corpus.

\citet{zabokrtsky-etal-2023-findings}'s evaluation includes two English-language datasets: GUM~\citep{10.1007/s10579-016-9343-x} and English ParCorFull \citep{lapshinova-koltunski-etal-2022-parcorfull2}. Of these we select GUM, as ParCorFull consists of only 19 documents. For GUM, we specifically use the OntoGUM formatted version of the dataset from \citet{zhu-etal-2021-ontogum} which allows us to consider the case where two datasets are intended to share definitions of the same theoretical construct, but differ in aspects related to their operationalization.

Furthermore, we include the recently released, large-scale Phrase Detectives 3.0 dataset~\citep{yu-etal-2023-aggregating} as a training set.

\subsubsection{Dataset Scope}

While \textit{coreference} is most often used to refer to \textit{identity coreference}, a relationship between linguistic expressions that refer to discourse entities with the same identity~\citep{nedoluzhko2021coreference}, there are many related phenomena which are sometimes also referred to as coreference---and may be considered instances of identity coreference in certain contexts---such as bridging anaphora~\citep{clark1977bridging,roesiger-etal-2018-bridging}, discourse deixis~\citep{webber1991structure,Zinsmeister2010TowardsAS},  event coreference (sometimes considered a subtype of discourse deixis)~\citep{cybulska-vossen-2014-using,lu2018event}, and split-antecedent anaphora~\citep{edes1968output}.
We focus on datasets of identity coreference and consider these additional types only insofar as they are included within the annotations of a dataset being studied.

Furthermore, as in \citet{xia-van-durme-2021-moving}, we exclude discontinuous mention spans~\citep{yu-etal-2023-universal} from our evaluations. Discontinuous spans are incompatible with most existing CR models which rely on the assumption that mentions correspond to a continuous span.

The phenomenon of \textit{singletons} is also sometimes considered related to identity coreference, although definitions of singletons differ. \citet{kubler-zhekova-2011-singletons} describe singletons as ``a cover term for mentions that are never coreferent, such as in \textit{in general} or \textit{on the contrary}, and mentions that are potentially coreferent but occur only once in a document.'' However, another common definition of a singleton is a linguistic expression that refers to a discourse entity which is referenced only once~\citep{can_we_fix}. The existence of singletons is one aspect of the definition of coreference that has been considered in recent multi-dataset evaluations \citep{toshniwal-etal-2021-generalization}. We focus our evaluation on non-singleton mentions, a common practice for multi-dataset evaluations~\citep{zabokrtsky-etal-2022-findings}.

\subsubsection{Dataset Formatting}

For PreCo, we use the last 500 documents of the training set as a validation split. For Phrase Detectives we randomly select 45 documents from the training set as a validation split. For Litbank we use ``split\_0'' from the official repository. For OntoGUM we use the official splits and include Reddit data. For ARRAU, we use the existing splits for the WSJ data and randomly split the remaining data into train/validation/test splits of size 80/10/10 percent, respectively.

\subsection{Detailed Constructs}
\label{sec:detailed-constructs}

\begin{table}[t]
    \centering
    \scriptsize
    \def\arraystretch{0.65}
    \begin{tabular}{lcccc}
        Dataset & Generic Only & VPs & Appositives & Copulae \\
        \midrule
        OntoNotes & \xmark & \cmark & \xmark\asterisk & \xmark \\
        PreCo & \cmark & \xmark & \cmark & \cmark \\
        Phrase Det. & \cmark & \xmark & \xmark\asterisk & \xmark\asterisk \\
        \bottomrule
    \end{tabular}
    \vspace{-6pt}
    \caption{Differences in definitions of \textit{coreference} as a theoretical construct (*annotated in the dataset, but not considered coreference).}
    \label{tab:difference-in-defs}
    \vspace{-10pt}
\end{table}

Differences in constructs are presented in Table~\ref{tab:difference-in-defs}. We provide additional details below.

\para{OntoNotes}
\textit{Identical coreference} is defined in the OntoNotes annotation guidelines to be ``names, nominal mentions, pronominal mentions, and verbal mentions of the same entity, concept, or event''~\citep{bbn2007guidelines}. This construct is distinguished from appositives (``immediately-adjacent noun phrases, separated
only by a comma, colon, dash, or parenthesis'') and copular structures (a subject and predicate linked by a copula). Furthermore, identical coreference is defined to ``not include entities that are only mentioned as generic, underspecified or abstract.''

\para{PreCo}
The PreCo authors establish a conceptualization of coreference mostly by contrasting what is considered to be coreferring in PreCo with the OntoNotes definition. As the PreCo annotation guidelines are not public, we can only assess the construct intended to be measured based on the corresponding paper and data. The PreCo authors note that they follow most of the conceptualization of coreference used in OntoNotes with the exception that they exclude verb phrases and explicitly include generic mentions and mentions in appositive and copular structures which may refer to the same entity. The PreCo dataset is additionally annotated for singleton mentions, although no definition is given for the concept of singleton.

\para{Phrase Detectives}
In the case of Phrase Detectives, coreference is considered a relation strictly between noun phrases. Predication, including apposition and copular structures, is distinguished from coreference and annotated separately. Generic only mentions can be annotated as coreferring.

\section{Model Hyperparameters}
\label{sec:model-hps}

For all models we  use the default hyperparameters, except that we use bf16 precision for training of LingMess models in order to decrease the total training time required.

\section{Statistical Significance}
\label{sec:statistical-significance}

We perform statistical significance tests using the same procedure as the MUC evaluation as described by \citet{chinchor-1992-statistical}. That is, we conduct a permutation test with 10,000 permutations and an $\alpha$ of 0.1. We use this significance test to highlight F1 scores where the difference relative to the in-domain model is significantly larger than the same difference calculated for all mentions.

\section{Additional Results}
\label{sec:addt-results}

\subsection{Additional Types}

While we do not focus on these types in the body of the paper, the noted differences in the annotation of verb phrases (VPs) and apposition can also be heuristically approximated for which we observe trends similar to the other types.

LingMess\textsubscript{PC} and LingMess\textsubscript{PD} never predict VPs as coreferring in the OntoNotes test set (where a VP is taken to be a mention that consists of only V* POS tags) compared with 15.2 / 100.0 / 26.3 disaggregated R / P / F1 for LingMess\textsubscript{ON}.

Appositives are annotated in OntoNotes and Phrase Detectives, but not considered identical coreference. If we approximate apposition as two coreferring mentions in the same sentence that are either adjacent or separated only by a single punctuation token, LingMess\textsubscript{ON} achieves 67.0 disaggregated F1 on appositives as compared to 13.3 and 25.0 for LingMess\textsubscript{PC} and LingMess\textsubscript{PD}, respectively, again showing a systematic drop in performance.

\vspace{2em}

\subsection{Generalization Gaps}

\begin{table}[!ht]
    \small
    \centering
    \begin{tabular}{|l|l|l|}
    \toprule
        & LingMess\textsubscript{PC} & LingMess\textsubscript{PD} \\
        \midrule
        OntoNotes & 26.6 & 41.8 \\
        \quad Nested & 52.0 & 55.4 \\
        \quad ON Generic & 40.1 & 34.4 \\
        \quad Compound & 57.1 & 81.0 \\
        \quad Copular & 49.2 & 50.0 \\
        \midrule
        OntoGUM & 14.3 & 22.7 \\
        \quad Nested & 0.0 & 0.0 \\
        \quad ON Generic & 13.1 & 14.4 \\
        \quad Compound & 26.3 & 47.5 \\
        \bottomrule
    \end{tabular}
    \caption{
    The AGG and TGG values calculated for Table~\ref{tab:results-ontonotes}. (I.e., the difference in the F1 score of the out-of-domain model as compared to the in-domain model.)
    }
\end{table}

\begin{table}[!ht]
    \small
    \centering
    \begin{tabular}{|l|l|l|}
    \toprule
        & LingMess\textsubscript{PC} & LingMess\textsubscript{PD} \\
        \midrule
        PreCo & 12.1 & 16.9 \\
        \quad Nested & 44.3 & 11.1 \\
        \quad ON Generic & 58.5 & 38.9 \\
        \quad Compound & 72.1 & 5.0 \\
        \quad Copular & 62.9 & 0.8 \\
        \midrule
        Phrase Det. & 12.9 & 33.0 \\
        \quad Nested & 24.5 & 23.9 \\
        \quad ON Generic & 34.6 & 36.1 \\
        \quad Compound & 18.5 & 8.1 \\
        \quad Copular & 0.7 & 0.0 \\
        \midrule
        LitBank & 19.3 & 33.4 \\
        \quad Nested & 17 & 24.3 \\
        \quad ON Generic & 10.8 & 7.5 \\
        \quad Compound & 0.0 & 0.0 \\
        \quad Copular & 21.9 & 28.6 \\
        \midrule
        ARRAU & 6.6 & 24.6 \\
        \quad Nested & 5.7 & 5.9 \\
        \quad ON Generic & 41.1 & 32.9 \\
        \quad Compound & 19.1 & 36.2 \\
        \quad Copular & 10.8 & 0.8 \\
        \midrule
        MMC & 6.3 & 16.8 \\
        \quad Nested & 1.0 & 10.0 \\
        \quad ON Generic & 36.4 & 18.2 \\
        \quad Compound & 10.3 & 100.0 \\
        \quad Copular & 53.9 & 0.0 \\
        \bottomrule
    \end{tabular}
    \caption{
    The AGG and TGG values calculated for Table~\ref{tab:results-all}. (I.e., the difference in the F1 score of the out-of-domain model as compared to the in-domain model.)
    }
\end{table}

\end{document}